\pdfoutput=1

\documentclass[11pt]{article}
\usepackage{booktabs}
\newcommand{\bftab}{\fontseries{b}\selectfont}
\newcommand{\under}{\underline}
\usepackage{multicol}
\usepackage{multirow}
\usepackage{graphicx}
\usepackage{acl}

\usepackage{times}
\usepackage{latexsym}

\usepackage[T1]{fontenc}

\usepackage[utf8]{inputenc}

\usepackage{microtype}

\usepackage{inconsolata}
\usepackage{import}
\usepackage{microtype}
\usepackage{layout}
\usepackage{tabularx, makecell}
\usepackage{booktabs}
\usepackage{mathrsfs}
\usepackage{amssymb} 
\usepackage{url}
\usepackage{graphicx}
\usepackage{xspace}
\usepackage{times,latexsym}
\usepackage{amsmath}
\usepackage{appendix}
\usepackage{comment} 
\usepackage{enumitem}
\usepackage{makecell}
\usepackage{multirow}
\usepackage{xcolor}
\usepackage{graphicx}
\usepackage{cleveref}
\usepackage{tcolorbox}
\usepackage{adjustbox}
\usepackage{relsize}
\usepackage{listings}
\usepackage{spverbatim}
\usepackage{tabularx}

\definecolor{mycolor}{HTML}{2650CC}

\newcommand{\z}{\phantom{0}}

\newcommand{\ex}[1]{\textit{#1}\xspace}

\newcommand{\tabref}[1]{Table~\ref{#1}\xspace}
\newcommand{\figref}[1]{Figure~\ref{#1}\xspace}

\newcommand{\appref}[1]{Appendix~\ref{#1}\xspace}

\title{Demystifying Instruction Mixing for Fine-tuning Large Language Models}

\author{
Renxi Wang\textsuperscript{1,2} \quad \textbf{Haonan Li\textsuperscript{1,2}} \quad Minghao Wu\textsuperscript{3} \quad Yuxia Wang\textsuperscript{1,2}
\\ \textbf{Xudong Han\textsuperscript{1,2}} \quad
\textbf{Chiyu Zhang\textsuperscript{4}} \quad \textbf{Timothy Baldwin\textsuperscript{1,2,5}}
\\ 
\textsuperscript{1}Mohamed bin Zayed University of Artificial Intelligence \quad \textsuperscript{2}LibrAI \\ 
\textsuperscript{3}Monash University \quad 
\textsuperscript{4} University of British Columbia \quad 
\textsuperscript{5} The University of Melbourne 
\\ \texttt{\{renxi.wang,haonan.li,yuxia.wang,xudong.han,timothy.baldwin\}@mbzuai.ac.ae} \\ \texttt{minghao.wu@monash.edu} \quad \texttt{chiyuzh@mail.ubc.ca}
}

\begin{document}
\maketitle

\begin{abstract}
Instruction tuning significantly enhances the performance of large language models (LLMs) across various tasks. However, the procedure to optimizing the mixing of instruction datasets for LLM fine-tuning is still poorly understood. This study categorizes instructions into three primary types: NLP downstream tasks, coding, and general chat. We explore the effects of instruction tuning on different combinations of datasets on LLM performance, and find that certain instruction types are more advantageous for specific applications but can negatively impact other areas.  This work provides insights into instruction mixtures, laying the foundations for future research.\footnote{Code and data are available at: \url{https://github.com/Reason-Wang/InstructLLM}.}
\end{abstract}

\section{Introduction}
Instruction tuning has been shown to have surprising efficacy for aligning large language models (LLMs) with human instructions \citep{chung2022scaling, li2023bactrianx, wu2023laminilm, xu2023wizardlm, touvron2023llama, muennighoff-etal-2023-crosslingual, gunasekar2023textbooks}. Recent studies highlight the diverse ways in which instructions can enhance the different capabilities of LLMs. For instance, using general-purpose, chat-like instructions can improve the performance of LLMs as
chat assistants \citep{vicuna2023, ouyang2022training, alpaca, ding2023enhancing}, while training LLMs on instructions based off NLP tasks improves their performance on NLP benchmarks \citep{sanh2022multitask, chung2022scaling, muennighoff2023crosslingual}, and incorporating coding instructions enhances LLM code generation \citep{fu2022gptroadmap, gunasekar2023textbooks}. However, a key unresolved issue is determining how to combine various instruction datasets to optimize overall LLM performance.

\begin{figure}
    \centering
    \includegraphics[width=\linewidth]{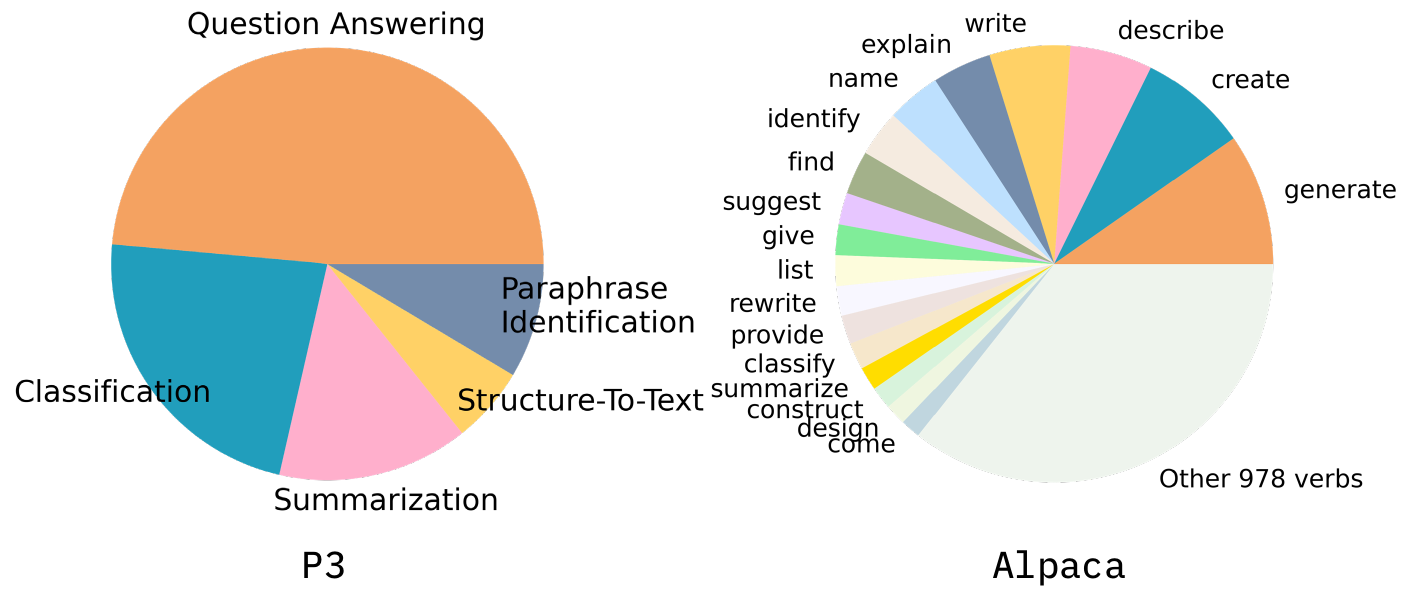}
    \caption{Instruction type distribution of P3 and Alpaca. For P3, the statistics come from the original dataset, while for Alpaca, we use a dependency parsing approach to extract the root verb of each instruction.}
    \label{fig:p3_alpaca}
\end{figure}

In this paper, we aim to better understand the impact of instruction mixing across three critical areas: NLP downstream tasks, coding, and chat. The core of our investigation revolves around understanding the influence of instruction dataset distributions on model performance in these different areas. We first select representative instruction datasets: P3 \citep{sanh2022multitask} for NLP downstream tasks, CodeAlpaca \citep{codealpaca} for code generation, and Alpaca \citep{alpaca} for general-purpose instructions. As shown in \figref{fig:p3_alpaca}, P3 is focused primarily on five tasks (including QA and classification), whereas Alpaca contains a vast array of instructions. Using a dependency parser, we identify over 1K unique root verbs from Alpaca's instructions, with \ex{generate}, \ex{create}, and \ex{describe} being the most frequent. CodeAlpaca, by contrast, is exclusively focused on coding tasks, and exhibits less variation compared to the others as exemplified in \tabref{tab:examples}. We fine-tune models across all eight potential combinations of these instruction datasets, and carry out detailed evaluation of model performance in terms of NLP downstream tasks, coding proficiency, and chat capabilities.


Our main contribution in this work is to shed light on instruction mixing when fine-tuning LLMs through comprehensive experimentation. Our findings can be summarized as follows:
\begin{itemize}
    \item Specific instruction datasets enhance LLM performance in their respective task areas. However, combining all instruction types does not uniformly improve performance across all tasks.
    \item Instructions reformulated from NLP downstream tasks (such as P3) can negatively impact the model's conversational abilities. In contrast, instructions focused on coding not only improve coding proficiency but also enhance chat capabilities.
    \item Larger models, with their increased capacity, are able to make more effective use of a diverse range of instructions.
\end{itemize}

\section{Related Work}

Recent work has demonstrated that vanilla LLMs can follow general instructions if tuned with instructions and corresponding responses \cite{mishra-etal-2022-cross, sanh2022multitask, wang-etal-2022-super}. 
For instance, \citet{sanh2022multitask} crafted an instructional dataset by reformulating supervised datasets with various prompts to create P3. However, despite their effectiveness in NLP tasks, these LLMs often diverge from human-like interactions in chatbot applications.

To facilitate general-purpose LLM fine-tuning, researchers has create general-purpose instructional data by human annotation \citep{DatabricksBlog2023DollyV2} and automatic approaches \citep{wang-etal-2023-self-instruct,alpaca}.
Recent work has further expanded the dataset size \citep{wu2023laminilm}, language coverage \citep{li2023bactrianx}, and task types \citep{codealpaca, yue2023mammoth}.

With increasing capabilities of LLMs and availability of instruction datasets, researchers have aimed to imbue a single model with diverse capabilities. 
\citet{sengupta2023jais} attempted to blend different instruction datasets without considering the data volume and task types. \citet{longpre2023flan} suggested that increasing the number of tasks and instruction diversity can enhance performance. 
In contrast, \citet{gpt4all} excluded P3 from their fine-tuning dataset, seemingly to enhance alignment. 
Nevertheless, none of these papers systematically studied the impact of the instruction mixture on the resulting LLM.

Concurrent to our work, \citet{wang2023far} fine-tuned LLaMA models on 12 instruction-tuning datasets separately. By evaluating those model on 7 tasks, they found that different datasets can enhance model performance on individual tasks. 
They further identified the optimal dataset combination, and trained a single model to achieve the best overall performance.
Novel to this work, we classify the instructions and model skills into three types, and conduct a deep analysis of the influence of data mixture on the models.

\section{Experimental Setup}\label{sec:expertimental_setup}

\begin{table*}[t]
\small
\centering
\begin{tabular}{llcccccccccc}
\toprule
\multirow{2}{*}{\textbf{Model}} & \multirow{2}{*}{\textbf{Data}} & \textbf{ARC} & \textbf{Wino-} & \multirow{2}{*}{\textbf{PIQA}} & \multirow{2}{*}{\textbf{MMLU}} & \multirow{2}{*}{\textbf{Race}} & \textbf{Hella-}& \multirow{2}{*}{\textbf{Average}} && \multicolumn{2}{c}{\textbf{HumanEval}}\\
                                & & \textbf{(challenge)} & \textbf{grande} & & & & \textbf{Swag} &&& \textbf{@1} & \textbf{@10}\\
\midrule
    \multirow{8}{*}{\textbf{LLaMA-2-7B}}  
    & None  & 43.1        & 69.5        & 78.0        & 40.8        & 39.2        & 57.2        & 54.6        && 13.7        & 21.3      \\
    & A     & 47.8        & 67.6        & 78.2        & 42.2        & \under{44.5}& \bftab 61.1 & 56.9        && 13.5        & 17.1       \\
    & C     & 46.1        & 69.5        & \under{78.5} & 41.0        & 41.1        & \under{61.0}& 56.2       && 16.2        & \under{24.4}\\
    & P     & \under{49.6}& \bftab 71.4 & \bftab 79.0 & \bftab 46.0 & 43.5        & 59.4        & \bftab 58.2 && \z4.6         & \z7.9        \\
    & AC    & 47.1        & 66.9        & 78.1        & 40.4        & 44.2        & 59.7        & 56.1        && \bftab 17.5 & \bftab 25.0\\
    & AP    & 48.4        & 70.0        & 78.1        & 43.8        & 42.9        & 58.5        & 56.9        && 13.8        & 17.7       \\
    & CP    & 48.0        & \under{71.3}& 78.4        & \under{44.9}& 44.4        & 60.7        & \under{57.9}&& \under{16.8}& 20.1       \\
    & ACP   & \bftab 49.7 & 68.0        & 77.9        & 43.5        & \bftab 44.6 & 58.7        & 57.1        && 16.0        & 23.8       \\
\midrule                                                                                                      
    \multirow{8}{*}{\textbf{LLaMA-2-13B}}                                                                     
    & None  & 48.6        & 71.9        & 79.2        & \bftab 52.1 & 40.7        & 60.1        & 58.8        && 15.4        & 26.2        \\
    & A     & 54.1        & 71.2        & 80.0        & 47.9        & \bftab 47.1 & \bftab 65.6 & 61.0        && 15.1        & 20.7        \\
    & C     & 49.7        & 73.4        & \bftab 80.8 & \under{51.5}& 45.4        & 63.6        & 60.7        && 17.9        & 24.4        \\
    & P     & 54.3        & \under{74.2}& 80.0        & 50.3        & \under{45.6}& 62.5        & \under{61.1}&& \z0.3       & \z1.8         \\
    & AC    & 51.6        & 68.8        & \under{80.6}& 48.7        & 44.4        & 63.0        & 59.5        && 17.1        & \under{27.4}\\
    & AP    & \under{54.8}& 71.7        & 80.3        & 51.2        & 45.2        & 62.7        & 61.0        &&  \z8.3        & 14.6        \\
    & CP    & \bftab 55.4 & \bftab 74.6 & 80.5        & 51.4        & \under{45.6}& \under{63.9}& \bftab 61.9 && \under{18.2}& 25.0        \\
    & ACP   & 54.4        & 71.5        & 80.0        & 50.0        & \bftab 47.1 & 63.1        & 61.0        && \bftab 20.2 & \bftab 32.9 \\
\bottomrule
\end{tabular}
\caption{Results on NLP and code generation benchmarks. All experiments are done in a zero-shot setting. The best result is in \textbf{bold}, and the second best result is \under{underlined}.} 
\label{tab:full_benchmarks}
\end{table*}

\paragraph{Datasets} 
We select Alpaca \citep{alpaca} as the \textit{general instruction dataset} to align models, in the form of 52K instruction--response pairs. 
We use P3 \citep{sanh2022multitask} as our \textit{NLP task instruction dataset}, which is reformatted for a wide range of NLP downstream tasks using diverse human-written templates. 
Since the number of samples in each task varies vastly, we randomly sample 1K instances from each subtask formatted with several corresponding prompts for diversity, resulting in 660K samples. 
For \textit{coding data}, we choose CodeAlpaca \cite{codealpaca}, which is an instruction dataset focusing on code generation. It contains 20K samples in different programming languages. 
To ensure a balanced comparison, we randomly sample a 20K subset from each dataset. Examples are provided in \tabref{tab:examples} in the Appendix.

\paragraph{Evaluation} 
We divide the evaluation into three parts: NLP benchmark performance, code generation, and alignment evaluation (i.e., chat ability evaluation). For NLP benchmarks, we use ARC \citep{clark2018think}, Winogrande \citep{sakaguchi2021winogrande}, PIQA \citep{bisk2020piqa}, MMLU \citep{hendrycks2020measuring},  RACE \citep{lai-etal-2017-race}, and HellaSwag \citep{zellers-etal-2019-hellaswag}. For coding, we use HumanEval \citep{chen2021evaluating}, which tests the pass rate of the generate codes. For alignment evaluation, we use the FLASK \citep{ye2023flask} framework to score model alignment. We keep the eight most frequent alignment skills from the original evaluation set, resulting in 1,180 samples. Then we employ GPT-4 to assess model responses to each instruction sample based on human-written principles. See Appendix \ref{sec:alignment_skills} for details of these skills.

\paragraph{Models}
We fine-tune LLaMA-2 7B and 13B \citep{touvron2023llama} models for two epochs in a generative way as in \citet{radford2018improving}, using a linear scheduler with a $3\%$ warmup rate and a batch size of 64. The maximum learning rate is $5\times10^{-5}$. 
The resources for training and evaluations are detailed in \appref{sec:resource}.

\begin{figure}[t]
    \centering
    \includegraphics[width=\linewidth]{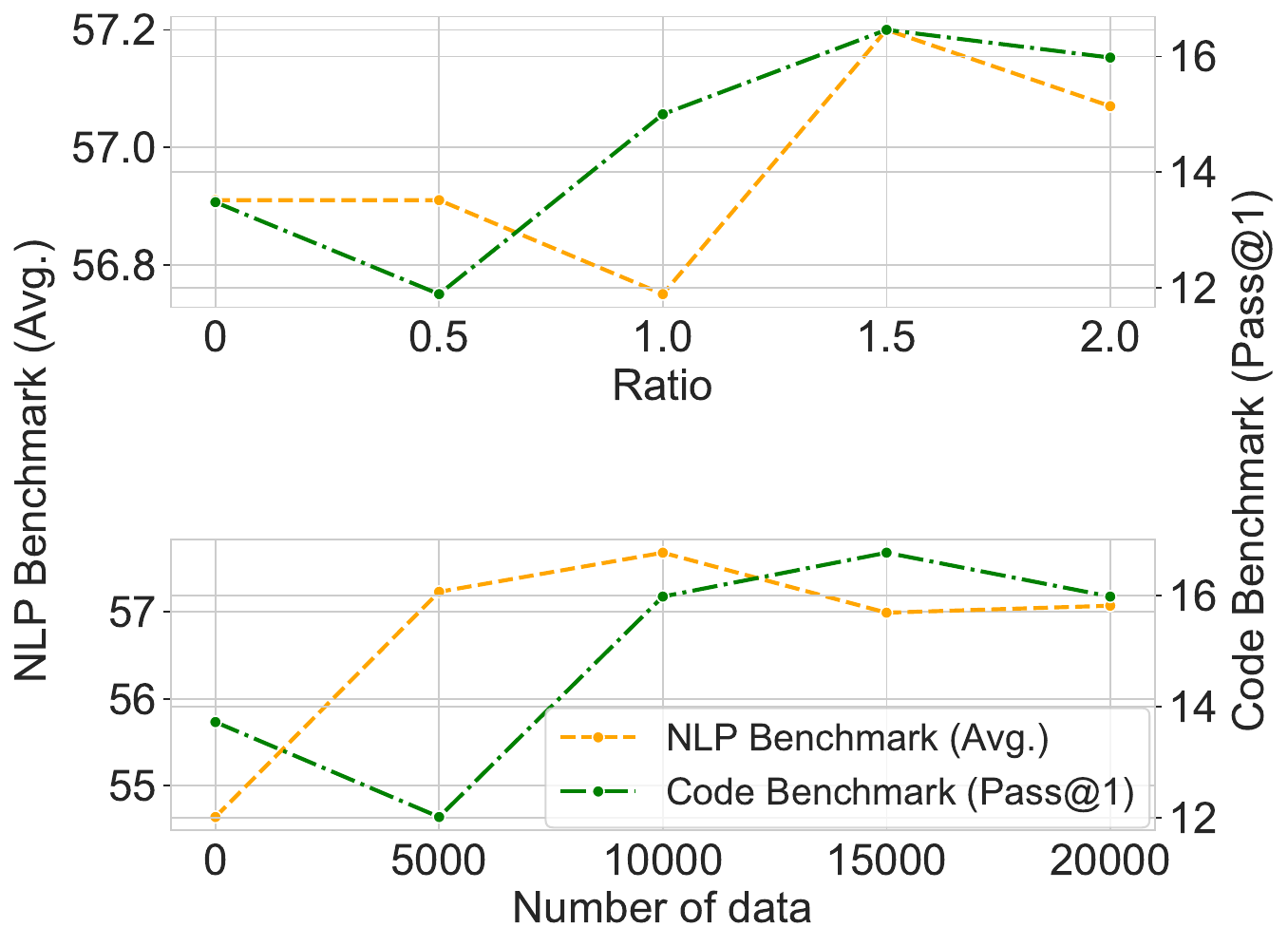}
    \caption{NLP benchmark scores (avg) and Code benchmark (HumanEval) scores for LLaMA-2-7B tuned with different mixing ratios and different numbers of instances. We keep the number of Alpaca instances constant at 20K and change the number of P3 and CodeAlpaca instances to get different ratios.}
    \label{fig:ratio_number}
\end{figure}

\section{Results}

For the remainder of this paper, we denote the Alpaca, CodeAlpaca, and P3 datasets as A, C, P, respectively. For each model, we compare eight different data mixing strategies, denoted as None, A, C, P, AC, AP, CP, ACP, where \ex{None} represents the vanilla model without fine-tuning, and each of the other settings represents the model fine-tuned with the corresponding dataset. For example, \ex{AC} means the model is fine-tuned with both Alpaca and CodeAlpaca. 

\begin{table*}[t]
\small
\centering
\begin{tabular}{llccccccccc}
\toprule
\textbf{Model} & \textbf{Data } & \textbf{Corr.} &  \textbf{Fact.} & \textbf{Comm.}  & \textbf{Compr.} & \textbf{Compl.} & \textbf{Insight.} & \textbf{Read.} & \textbf{Conc.} & \textbf{Avg.} \\
\midrule
    \multirow{7}{*}{\textbf{LLaMA-2-7B}} 
    & A    & 47.6        & \bftab 55.4 & 58.8        & \under{54.8}& \under{48.0}& \bftab 50.4 & \bftab 88.0 & 81.6        & \under{60.6}\\
    & C    & 48.8        & 52.0        & 58.4        & 52.0        & 40.2        & 46.2        & 83.8        & 78.4        & 57.4        \\
    & P    & 47.2        & 40.0        & 48.8        & 38.4        & 29.0        & 30.4        & 64.4        & 68.6        & 45.8        \\
    & AC   & \under{49.0}& \under{54.4}& \bftab 59.6 & \bftab 56.4 & \bftab 48.2 & \under{49.8}& \under{86.6}& \bftab 85.6 & \bftab 61.2 \\
    & AP   & 48.4        & 51.4        & 57.6        & 52.6        & 45.0        & 46.0        & 84.2        & 80.8        & 58.2        \\
    & CP   & 47.0        & 49.6        & 54.2        & 48.8        & 36.2        & 41.8        & 78.2        & 77.2        & 54.2        \\
    & ACP  & \bftab 50.4 & 53.0        & \under{59.0}& 53.8        & 47.2        & 46.8        & 85.0        & \under{81.8}& 59.6        \\
\midrule
    \multirow{7}{*}{\textbf{LLaMA-2-13B}}
    & A    & 53.6        & \under{58.8}& \under{63.8}& \under{60.0}& \under{47.6}& \bftab 55.2 & \bftab 89.2 & \under{84.0}& \under{64.0} \\
    & C    & \bftab 57.2 & \under{58.8}& 61.0        & 57.8        & 43.8        & 52.4        & 85.6        & 82.2        & 62.4         \\
    & P    & 49.4        & 42.4        & 51.8        & 42.0        & 28.2        & 32.0        & 66.8        & 70.4        & 47.8         \\
    & AC   & \under{55.6}& \bftab 61.0 & \bftab 66.6 & \bftab 61.2 & \bftab 51.4 & \under{54.0}& \under{88.4}& \bftab 86.6 & \bftab 65.6  \\
    & AP   & 53.0        & 55.4        & 60.6        & 56.2        & 47.0        & 48.0        & 85.0        & 83.4        & 61.0         \\
    & CP   & 53.0        & 53.2        & 57.4        & 53.4        & 39.0        & 45.2        & 81.2        & 82.6        & 58.2         \\
    & ACP  & 51.6        & 55.6        & 61.8        & 57.0        & 47.0        & 48.6        & 87.0        & 83.0        & 61.4         \\
\bottomrule
\end{tabular}
\caption{GPT-4 evaluation results on alignment skill assessment. We report eight dimensions: logical correctness, factuality, commonsense understanding, comprehension, completeness, insightfulness, readability, and conciseness, as well as average scores. Since the vanilla model cannot follow instructions, we exclude it from this table.
The best result is in \textbf{bold}, and the second best result is \under{underlined}.}   
\label{tab:full_skills}
\end{table*}

\subsection{NLP Tasks and Code Benchmark Results}

\tabref{tab:full_benchmarks} shows the zero-shot results on the NLP and code generation benchmarks. 
Predictably \emph{each specialized instruction dataset improves the performance on the benchmarks they are designed for}. 
In the no-mixture setting (comparing A, C, and P), models fine-tuned on P3 achieve the highest average score for NLP tasks, while models fine-tuned on CodeAlpaca excel in code generation benchmarks.
Examining specific tasks reveals that \emph{a model's performance on a specific task heavily relies on the similarity between the target task and the tasks it was fine-tuned on}. For instance, Alpaca fine-tuned models excel in Race and HellaSwag, which involve the story completion task, similar to the Alpaca instruction format. On the other hand, P3 fine-tuned models perform well on ARC and Winogrande, which involve multiple-choice QA and cloze tests, which are well represented in P3.

In the mixture setting, it's evident that \emph{including specialized data consistently boosts model performance in corresponding benchmarks compared to models without such data.} 
For example, P, PA, PC, and PCA perform better than None, A, C, and CA on NLP downstream tasks. 
Focusing on the code benchmarks, \emph{incorporating general instructions consistently improves coding performance}. For the 7B model, AC improves performance by $+$1.28 and $+$0.61 compared to C, while the improvements are $-$0.80 (outlier) and $+$3.05 for the 13B models. 
Another interesting finding is that the 13B models perform best with the ACP mixture, while the 7B models perform best with AC. This suggests that \emph{larger models can better learn from varied instructions more effectively than smaller models}.

These findings highlight the importance of considering model size and target usage when designing the instruction mixture.

\paragraph{Mixing with Different Ratios} 
While it is clear that mixing specialized instructions is vital for benchmark performance, how the mixing ratio correlates with the performance is also important.
As \figref{fig:ratio_number} shows, with the number of general instructions fixed to 20K, scores in both NLP task and code benchmarks first decrease and then increase as the ratio of specialized instructions increases. 
They both peak when the ratio is 1.5, and drop back slightly when the ratio is increased further to 2.0.
We hypothesize that this is because the model overfits to the specialized instructions when there are too many such instructions.

\paragraph{Number of instances} \figref{fig:ratio_number} also shows the performance change with respect to the number of fine-tuning data instances. 
We mix each type of instruction with the same number. 
We find that the performance over both benchmarks plateaus when the number of instances is larger than 10K.

\subsection{Alignment Skill Results}\label{sec:alignment_results}


\tabref{tab:full_skills} shows the alignment skills results. We adopt the same setup as FLASK, using \texttt{GPT-4-0613} to access the alignment skills and scaling the scores to the range $[0, 100]$.

From \tabref{tab:full_skills} we make the following observations: 
(1) \emph{All three types of instructions improve model alignment compared to the vanilla LLM.} Among these instructions, Alpaca stands out as the most effective. It contains general-purpose instructions and human-like responses, making it a better fit for aligning models with humans.
(2) \emph{While CodeAlpaca alone doesn't notably enhance alignment abilities, combining it with general instructions results in a substantial improvement of $+$0.6 (7B) and $+$1.6 (13B) points};
these improvements are mainly due to better compression, commonsense understanding, completeness, and conciseness.
(3) \emph{Mixing P3 data causes a drop of $-$2.8 (7B) and $-$3.6 (13B) in alignment skills,} 
suggesting that P3 has a negative impact on fine-tuning chatbot LLMs.

\section{Conclusion}
In this paper, we investigated different data mixing strategies in instruction fine-tuning. We measured models against diverse benchmarks and alignment skills. We find that general instructions provide better alignment as well as performance on NLP benchmarks, code instructions improve coding and alignment skills, while NLP task instructions hinder alignment skills when combined with other instruction types.

\section*{Limitations}
Our work is subject to several limitations that should be addressed in future research.
(1) We only use LLaMA-2 7B and 13B models in our experiments. Other models of varying sizes should be used to further verify our findings. We acknowledge that the model's behavior may vary with different sizes, and that usually, larger models have stronger capabilities, and hence may be able to handle more instructions without performance degradation.
(2) In this paper, we limit our instruction dataset to 20K and mainly compare the 1:1 ratio of all instruction types. We leave the exploration of the impact of more instructions and mixing ratios to future work.

We acknowledge these limitations and propose that future work should focus on addressing them to help the community better understand the impact of instruction mixture on LLMs.

\bibliography{anthology,custom}

\appendix

\section{Examples of Instruction Types}

\tabref{tab:examples} list examples from Alpaca, CodeAlpaca, and P3 dataset.
\begin{table*}[t]
\centering
\begin{tabularx}{\textwidth}{l|X}
\toprule
\textbf{Source} & \textbf{Example} \\
\midrule
Alpaca & \textbf{Instruction}: List 5 reasons why someone should learn to code. \\ & \textbf{Response}: 1. High demand for coding skills in the job market \textbackslash n2. Increased problem-solving and analytical skills \textbackslash  n3. Ability to develop new products and technologiesn \textbackslash n4. Potentially higher earning potential \textbackslash n5. Opportunity to work remotely and/or freelance\\
\midrule
    CodeAlpaca & \textbf{Instruction}: Write a function to find the number of distinct states in a given matrix. Input: matrix = [[1, 0, 0],\textbackslash n          [1, 0, 1],\textbackslash n          [1, 1, 1]] \\ & \textbf{Response}: def find\_num\_distinct\_states(matrix):\textbackslash n    states = set()\textbackslash n    for row in matrix:\textbackslash n        state = "".join([str(x) for x in row])\textbackslash n        states.add(state)\textbackslash n    return len(states)\\
\midrule
    P3 & \textbf{Instruction}: Answer the following question.\textbackslash nAnna Kournikova, Michelangelo, ILOVEYOU, Melissa, and Stuxnet are all examples of what? \\ &  \textbf{Response}: Computer virus/worm\\
\bottomrule
\end{tabularx}
\caption{Examples from Alpaca, CodeAlpaca, and P3.}   
\label{tab:examples}
\end{table*}

\section{Alignment Skills Demonstration}\label{sec:alignment_skills}
The FLASK framework annotates each instruction with three skills that is needed to respond to the instruction. We select 8 most frequent skills and filter out instructions annotated with other skills, resulting 1,180 instructions in the evaluation set. The following are demonstrations of each alignment skill from the annotation prompt.

\paragraph{Logical Correctness} Is the final answer provided by the response logically accurate and correct for an instruction that has a deterministic answer?

\paragraph{Factuality} Did the model extract pertinent and accurate background knowledge without any misinformation when factual knowledge retrieval is needed? Is the response supported by reliable evidence or citation of the source of its information?

\paragraph{Commonse Understanding} Is the model accurately interpreting world concepts for instructions that require a simulation of the expected result or necessitate commonsense or spatial reasoning?

\paragraph{Comprehension} Does the response fulfill the requirements of the instruction by providing relevant information especially when the instruction is complex and includes multiple requirements? This includes responding in accordance with the explicit and implicit purpose of given instruction.

\paragraph{Completeness} Does the response provide a sufficient explanation? Comprehensiveness and thoroughness of the response should be considered, which depends on the breadth of topics covered and the level of detail provided within each topic.

\paragraph{Insightfulness} Is the response creative, original or novel, including new perspectives or interpretations of existing information?

\paragraph{Readability} Is the response structured to promote readability and coherence? Does the response exhibit excellent organization?

\paragraph{Conciseness} Is the response presented in a concise manner for the reader without any unnecessary information?

The prompt for alignment skill assessment are provided in \figref{fig:flask_prompt}.
For how a response corresponds to a specific level of an alignment skill and other details, please refer to their repository.\footnote{https://github.com/kaistAI/FLASK}

\begin{figure*}[t]
\small
\begin{tcolorbox}[colframe=white, left=3mm, right=3mm]
[System]\\

We would like to request your feedback on the performance of the response of the assistant to the user instruction displayed below. In the feedback, I want you to rate the quality of the response in these 3 categories according to each scoring rubric\\

\textcolor{mycolor}{Skill 1 definition}

\textcolor{mycolor}{Skill 1 scoring principles}\\

\textcolor{mycolor}{Skill 2 definition}

\textcolor{mycolor}{Skill 2 scoring principles}\\

\textcolor{mycolor}{Skill 3 definition}

\textcolor{mycolor}{Skill 3 scoring principles}\\

[Instruction]

\textcolor{mycolor}{instruction}\\

[Ground Truth Answer]

\textcolor{mycolor}{ground truth answer}\\

[Assistant's Response]

\textcolor{mycolor}{response for evaluation}\\

[The End of Assistant's Response]\\

Please give feedback on the assistant's responses. Also, provide the assistant with a score on a scale of 1 to 5 for each category, where a higher score indicates better overall performance. Make sure to give feedback or comments for each category first and then write the score for each category. Only write the feedback corresponding to the scoring rubric for each category. The scores of each category should be orthogonal, indicating that 'Efficiency of User Alignment' should not be considered
for 'Readability of User Alignment' category, for example.\\

Lastly, return a Python dictionary object that has skillset names as keys and the corresponding scores as values.\\

[System]
\end{tcolorbox}
\caption{Alignment skill assessment prompt (from FLASK \citep{ye2023flask}). The blue parts are filled by corresponding content.}
\label{fig:flask_prompt}
\end{figure*}

\section{Resources}\label{sec:resource}
We use 4$\times$A100 to train LLaMA-2-7B and 8$\times$A100 to train LLaMA-2-13B. Each 20k data takes about 2 hours. For all experiments, training takes about 288 A100 GPU hours.

For evaluation, we use GPT-4, where each input has an average of 950 tokens and each output has an average of 293 tokens. All evaluations cost about \$760.

\end{document}